\documentclass[journal]{IEEEtran}
\ifCLASSINFOpdf
\else
\fi
\usepackage{amssymb}
\usepackage{graphicx}
\usepackage{amsmath}
\usepackage{algorithm}
\usepackage{algorithmic}
\usepackage{enumerate}
\usepackage{color,xcolor}
\usepackage{subfigure}
\usepackage{footnote}
\usepackage{url}
\usepackage[normalem]{ulem}  
\usepackage{bm}
\usepackage{float}
\usepackage{graphicx,times,amsmath,multirow,booktabs}
\usepackage{tabularx,booktabs}
\newcolumntype{Y}{>{\centering\arraybackslash}X}
\usepackage[colorlinks=true, linktocpage=true, linkcolor=blue, anchorcolor=blue, citecolor=green]{hyperref}

\makeatletter

\newcommand{\Rmnum}[1]{\expandafter\@slowromancap\romannumeral #1@}

\makeatother

\begin{document}
%
\title{Competitive Coevolution as an Adversarial Approach to Dynamic Optimization}
%
%
%

\author{Xiaofen~Lu, Ke~Tang,~\IEEEmembership{Senior Member,~IEEE}, Stefan~Menzel,
        and~Xin~Yao,~\IEEEmembership{Fellow,~IEEE}
\thanks{X. Lu, K. Tang (\textit{corresponding author}) and X. Yao are with Shenzhen Key Laboratory of Computational Intelligence, University Key Laboratory of Evolving Intelligent Systems of Guangdong Province, Department of Computer Science and Engineering, Southern University of Science and Technology, Shenzhen 518055, China. S. Menzel is with Honda Research Institute Europe, 63073 Offenbach/Main, Germany. X. Lu and X. Yao are also with the Centre of Excellence for Research in Computational Intelligence and Applications (CERCIA), School of Computer Science, University of Birmingham , Edgbaston, Birmingham B15 2TT, U. K. \textit{E-mails}: luxf@sustech.edu.cn, tangk3@sustech.edu.cn, stefan.menzel@honda-ri.de, xiny@sustech.edu.cn}}
\maketitle

\begin{abstract}
Dynamic optimization, for which the objective functions change over time, has attracted intensive investigations due to the inherent uncertainty associated with many real-world problems. For its robustness with respect to noise, Evolutionary Algorithms (EAs) have been expected to have great potential for dynamic optimization. On the other hand, EAs are also criticized for its high computational complexity, which appears to be contradictory to the core requirement of real-world dynamic optimization, i.e., fast adaptation (typically in terms of wall-clock time) to the environmental change. So far, whether EAs would indeed lead to a truly effective approach for real-world dynamic optimization remain unclear. In this paper, a new framework of employing EAs in the context of dynamic optimization is explored. We suggest that, instead of online evolving (searching) solutions for the ever-changing objective function, EAs are more suitable for acquiring an archive of solutions in an offline way, which could be adopted to construct a system to provide high-quality solutions efficiently in a dynamic environment. To be specific, we first re-formulate dynamic optimization problems as static set-oriented optimization problems. Then, a particular type of EAs, namely competitive coevolution, is employed to search for the archive of solutions in an adversarial way. The general framework is instantiated for continuous dynamic constrained optimization problems, and the empirical results showed the potential of the proposed framework. 
\end{abstract}


%
\IEEEpeerreviewmaketitle

\section{Introduction}
%
%
%
%


\IEEEPARstart{D}{ynamic} optimization problems (DOPs) widely exist in the real world due to the dynamic environment. For DOPs, their objective functions, constraints or even decision variables change over time. This leads to the change of their global optima over time. For example, in the real-time vehicle routing problems, the number of vehicles, the vehicle load or speed will change over time due to the wear of vehicles, meanwhile the customers' requirements or the traffic condition also change over time, and these will result in the constant change of the best route \cite{ghiani2003real}. The challenge of DOPs exists in that an optimizer needs to respond to ever-changing environment, i.e. quickly finding the new optimal solution once the problem changes \cite{nguyen2012evolutionary}. 


As a class of nature-inspired optimisation methods, evolutionary algorithms (EAs) have shown good robustness with respect to noise, and thus have been widely studied in the field of dynamic optimization. In the literature, many dynamic optimization methods have been developed based on EAs. Generally, these methods can be grouped into three categories: diversity-driven methods, memory methods and prediction methods \cite{nguyen2012evolutionary}. Prediction methods were proposed for predictable DOPs that change following certain rules. They predict new optimal solutions based on the best solutions obtained before and use them for population re-initialization once the problem changes \cite{hatzakis2006dynamic, rossi2008tracking, filipiak2014infeasibility}. Diversity-driven and memory methods are appropriate for unpredictable DOPs. Diversity-driven methods mainly include introducing diversity after a change happens \cite{cobb1990investigation, nguyen2012continuous}, maintaining diversity during search \cite{grefenstette1992genetic,campos2014bare}, and multi-population methods that maintain multiple subpopulations to locate and track multiple optima simultaneously \cite{li2015multi, bu2017continuous}. Memory methods store and use good-performing solutions in previous search \cite{richter2010memory} . 

In this paper, we focus on using EAs to solve unpredictable DOPs. For unpredictable DOPs, the logic behind diversity-driven and memory methods is to obtain some good solutions that are closer to the new optimal solution after change by introducing or maintaining diversity or using previously good-performing solutions, and thus achieve fast response to changes. However, for memory methods, the effectiveness of the logic actually depends on the assumption that the new problem after change is similar to a previously met problem, which might be invalid especially in a drastically-changing environment. Moreover, since memory methods depend on good-performing solutions obtained in previous search, they will suffer from cold-start problems. For diversity-driven methods, randomly introducing individuals can not guarantee obtaining good solutions except that the size is big enough, and thus are inefficient. Multi-population methods also have an assumption about the moving-peak characteristic of the DOPs that are to solve. Therefore, new dynamic optimization methods are still needed for unpredictable DOPs. 

In many real-world applications, fast online adaption to environmental changes is one of the most desirable features of an optimization algorithm for DOPs. For example, a vehicle routing strategy in logistics must respond quickly to the traffic condition or it might not satisfy users' requirements \cite{ghiani2003real}. Also, a high-frequency trading system needs to continually make an efficient decision to deal with rapid changes in financial asset price \cite{filipiak2017dynamic}. In this context, suppose we can maintain an archive in advance that contains some candidate solutions that are close to the optimal solution(s) after each change, then searching starting from this archive after each change will generally result in faster adaptation than diversity-driven methods. This is because conducting local search from one of these candidates can easily find the new optimal solution. Moreover, as the archive is obtained in the preparation stage before the whole system runs, the cold-start problem can be avoided compared to memory methods. The strategy described above is actually a system construction methodology that is different from the existing evolutionary dynamic optimization. This methodology may require additional computational cost to find a set of good candidate solutions, but it can meet the core requirement of real-world DOPs. In a sense, this might be a little analogous to neural network learning, where training could be very time-consuming and take a very long time, but testing of a trained neural network is very fast. 

The difficulty in the implementation of this methodology exists in how to find such a solution set. Ideally, if the optimal solution of each change for a DOP can be found beforehand, then they can form the archive. However, the information about each change is hard to know beforehand. Another way to obtain the solution set is to randomly sample a big set of environmental changes and find the optimal solution for each sampled change. However, the efficacy would be very low. 

Considering these, we propose in this paper a new dynamic optimization framework based on EAs which employs competitive co-evolution \cite{rosin1997new}, a particular type of EAs, to offline search an archive of good solutions in an adversarial way in the system design phase, and conducts local search on this archive to do fast online optimization for DOPs when the system runs. The general framework is instantiated for continuous dynamic constrained optimization problems (DCOPs), and the empirical results showed the potential of the proposed framework. 

The rest of this paper is organized as follows. Section \Rmnum{2} will present the problem formulation and explain why competitive co-evolution is chosen. Section \Rmnum{3} will give the proposed dynamic optimization framework and detail an instantiation of the framework for DCOPs. In Section \Rmnum{4}, the experimental results and analysis on dynamic constrained optimisation benchmark problems will be given. Section \Rmnum{5} will conclude this paper and point out future work directions.

\section{Problem Formulation}
Without loss of generality, the DOPs considered in this paper has the following formulation:
\begin{equation}\label{eq1}
\underset{\bm{x}} {\textrm{min}}\ f(\bm{x},\bm{\alpha}(t)) 
\end{equation}
where $\bm{x}=[x_{0},x_{1},...,x_{D_{x}}]^{T}$ denotes the decision variable vector and $D_{x}$ is the dimension of the search space, $\bm{\alpha}(t)=[\alpha_{0}(t),\alpha_{1}(t),...,\alpha_{D_{\alpha}}(t)]$ is the vector of environmental parameters which vary at a certain frequency as time goes by and $D_{\alpha}$ is the number of environmental parameters. In this work, for brevity, we assume $D_{x}$, $D_{\alpha}$, the search space of $\bm{x}$ and the variation space of $\bm{\alpha}$ are fixed. Also, the search space of $\bm{x}$ and the variation space of $\bm{\alpha}(t)$ are assumed to be known beforehand. Actually, this work can be extended to DOPs with $D_{x}$, $D_{\alpha}$, the search space of $\bm{x}$ and the variation space of $\bm{\alpha}(t)$ changing over time. 


As mentioned before, this work aims to solve the dynamic problem defined in Eq.~(\ref{eq1}) by offline finding a solution set that contains some individuals that are close to the optimal solution(s) for each problem change and then using these individuals as starting points for fast online optimization when the environment changes. Starting from this, the problem to find such a solution set can be formulated as follows:
\begin{equation}\label{eq2}
\begin{split}
\underset{\bm{X}} {\textrm{min}}\ & F(\bm{X},\bm{\alpha}(t)), \forall t \\
\textrm{where} \ & F(\bm{X},\bm{\alpha}(t))= \underset{\bm{x}\in\bm{X}} {\textrm{min}} f(\bm{x},\bm{\alpha}(t))\\
\end{split}
\end{equation}
Here $\bm{X}=\{\bm{x}_1,\bm{x}_2, ...., \bm{x}_\mathit{setsize}\}$ denotes the solution set to find and $setsize$ denotes the number of solutions in this set. 




It can be seen through Eq.~(\ref{eq2}) that the dynamic problem defined in Eq.~(\ref{eq1}) is re-formulated as a static set-oriented optimization problem. For Eq.~(\ref{eq2}), if we can know the value of $\bm{\alpha}(t)$ at each $t$, then $\bm{X}$ can be formed by the optimal solution for each $f(\bm{x},\bm{\alpha}(t))$. However, the value of $\bm{\alpha}(t)$ at each $t$ is hard to know beforehand. Instead, we can handle the problem in Eq.~(\ref{eq2}) by solving the following optimization problem:
\begin{equation}\label{eq3}
\begin{split}
\underset{\bm{X}} {\textrm{min}}\ & F(\bm{X},\bm{\alpha}), \forall \bm{\alpha} \\
\textrm{where} \ & F(\bm{X},\bm{\alpha})= \underset{\bm{x}\in\bm{X}} {\textrm{min}} f(\bm{x},\bm{\alpha})\\
\end{split}
\end{equation}
In this problem, since every possible value of $\bm{\alpha}$ needs to be considered, the process is time intractable. Actually, the optimization problem in Eq. ~(\ref{eq3}) can be changed to:
\begin{equation}\label{eq4}
\begin{split}
\underset{\bm{X}} {\textrm{min}}\ & F(\bm{X},\bm{\alpha})- \underset{\bm{x}} {\textrm{min}}f(\bm{x},\bm{\alpha}), \forall \bm{\alpha} \\
\textrm{where} \ & F(\bm{X},\bm{\alpha})= \underset{\bm{x}\in\bm{X}} {\textrm{min}} f(\bm{x},\bm{\alpha})\\
\end{split}
\end{equation}
Furthermore, we can get the following minimax optimization problem from Eq. ~(\ref{eq4}) :
\begin{equation}\label{eq5}
\begin{split}
\underset{\bm{X}} {\textrm{min}} \ \underset{\bm{\alpha}} {\textrm{max}} &\ (F(\bm{X},\bm{\alpha})- \underset{\bm{x}} {\textrm{min}}f(\bm{x},\bm{\alpha})) \\
\textrm{where} \ & F(\bm{X},\bm{\alpha})= \underset{\bm{x}\in\bm{X}} {\textrm{min}} f(\bm{x},\bm{\alpha})\\
\end{split}
\end{equation}
For this minimax optimization problem, the evaluation of $\bm{X}$ can be quite time-consuming because of large number of evaluations if $\bm{\alpha}$ changes in a continuous space. 

In the field of evolutionary computation, competitive co-evolution (CC) \cite{rosin1997new} have been shown to be suitable for minimax black-box optimization \cite{al2018towards}. In \cite{hillis1990co}, competitive co-evolution (CC) has been successfully applied to evolve sorting networks and a test suite simultaneously. In \cite{herrmann1999genetic}, CC was used on a simple parallel machine scheduling problem by co-evolving the assignments of tasks to machines and the processing times of the tasks to minimize the worst case makespan of the schedule. In \cite{al2018towards}, CC was applied to training generative adversarial networks (GANs). Inspired from biology, CC co-evolves a prey population and a predator population in an adversarial way to solve problems without explicit objective function. During the evolutionary process, the fitness of each individual in the prey population depends on its performance on the individuals in the predator population and vice versa. Through this adversarial way, both of the two populations can be enhanced. 

Motivated from this, we propose in this paper to apply CC to solve the minimax optimization problem in Eq. ~(\ref{eq5}). The prey population is used to search the solution set $\bm{X}$ and the predator population is used to search the environment $\bm{\alpha}$. The two populations are co-evolved to find the solution set with the optimal performance with respect to the worst possible predator $\bm{\alpha}$. More details about the CC-based dynamic optimization framework are given in Section \Rmnum{3}. It should be noted here that it might be impossible that all DOPs can be solvable through optimizing the problem in Eq. ~(\ref{eq5}) since the general idea has the assumption that a sufficiently limited set of solutions could fulfill the need of most (if not all) environments that would be encountered in the future. In other words, the performance of the general idea should be problem-dependent. 


\section{The Proposed Approach}
In this section, the proposed EA-based dynamic optimization framework is first introduced. Then, a new dynamic optimization algorithm for DCOPs is detailed by instantiating the proposed framework on DCOPs. 

\subsection{The General Framework}

\begin{figure}[!htbp]
	\centering
	\includegraphics[width=3.4in]{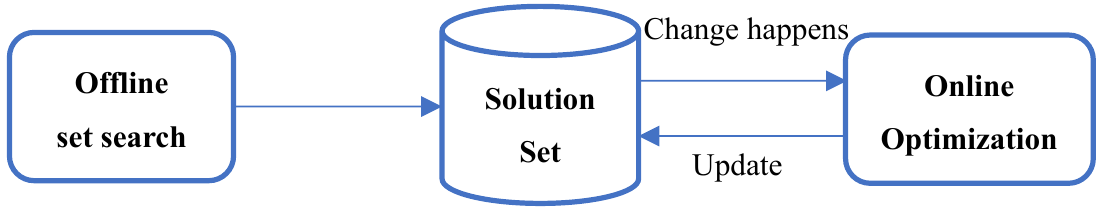}
	\caption{The New Dynamic Optimization Workflow}  \label{fig1}
\end{figure}

The workflow for the new dynamic optimization methodology is given in Figure~\ref{fig1}. The general idea is to offline search a solution set for a DOP that contains good solutions for different environmental parameter values before the whole system runs (i.e. system design phase), save it in an archive and then solve the DOP online based on the archive. In the online optimization phase, each time a problem change is detected, the online search is restarted to find a new solution based on the archive for the changed problem. Moreover, the solution set in the archive is also updated according to the newly encountered environmental change. 

Based on this idea, we propose a new EA-based dynamic optimization framework, which employs CC to offline search the solution set and online does local search on this solution set once a change is detected. Algorithm~\ref{alg1} gives the procedure of the new framework for dynamic optimization. In this algorithm, $D_{x}$ and $D_{a}$ are the dimension of the solution space and environment space for the DCOP, respectively. In the following subsections, both offline and online optimization parts will be detailed.

\begin{algorithm}[htbp]
	\algsetup{linenosize=\small} \small
	\caption{The Proposed EA-based Dynamic Optimization Framework} \label{alg1}
	\begin{algorithmic}[1]
		\REQUIRE ~~ \\
		The search space for the solution: $\bm{R}^{D_{x}}$  \\
		The range for the environmental parameters: $\bm{R^{D_{a}}}$  \\
		The maximum number of generations: $G_{\mathit{max}}$  \\
		The number of individuals to detect changes: $\mathit{detect}_{k}$ \\
		The memory to store locally best soluiton: $\mathit{mem}_\mathit{best}$ \\
		\STATE Do competitive co-evolution search based on $\bm{R}^{D_{x}}$ and $\bm{R^{D_{a}}}$ \\ for $G_{\mathit{max}}$ generations to get a solutions set $\bm{X}$
		\STATE Archive solution set $\bm{X}$ found by CC
		\STATE Initialize a population $\mathit{pop}$ based on $\bm{X}$
		\STATE Evaluate $\mathit{pop}$ with the current $f$
		\WHILE{stopping criteria is not satisfied}
		\STATE Do local search on $\mathit{pop}$ 
		\STATE Re-evaluate $\mathit{detect_{k}}$ sentinel solutions to detect changes
		\IF{change is detected}
		\STATE Update the solution set $\mathit{SP}_{G_\mathit{max}}$
		\STATE Re-intialize the population
		\ENDIF
		\ENDWHILE	
	\end{algorithmic}
\end{algorithm}

\subsubsection{Offline Set Search}
\begin{figure}[!htbp]
	\centering
	\includegraphics[width=1.8in]{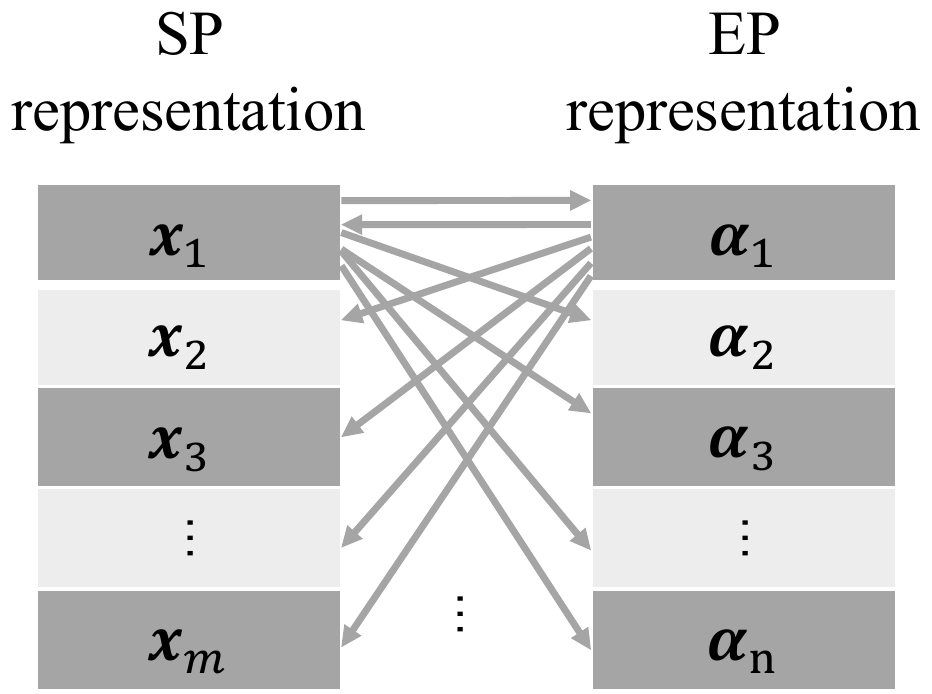}
	\caption{Representation for SP and EP}  \label{fig2}
\end{figure}

In the offline optimization phase, the framework applies CC to solve the minimax optimization problem in Eq.~(\ref{eq5}) to search for the solution set. The prey population is used to represent the solution set and the predator population to represent the environmental parameters. The prey population is denoted as SP (solution population) and the predator population is denoted as EP (environment population). Figure~\ref{fig2} shows the representation of SP and EP. In SP, each individual is denoted as $\bm{x}_{i}(i=1,2,...,m)$. In EP, each individual is denoted as $\bm{\alpha}_{j}(j=1,2,...,n)$. The size of SP and EP are recorded as $m$ and $n$, respectively. Each $\bm{x}_{i}$ is a candidate solution for the dynamic problem under the environment of $\bm{\alpha}_{j}$. The evaluation of each individual in SP (EP) depends on the individuals in EP (SP). The lines with an arrow are used to show such a testing and being tested relationship between SP and EP. 

Algorithm~\ref{alg5} gives the co-evolutionary framework for SP and EP. In this algorithm, $D_{x}$ and $D_{a}$ are the dimension of the solution space and environment space for the DCOP, respectively. At the beginning, CCDO intialises both SP and EP by randomly generating $m$ individuals from the search space of $\bm{x}$ and $n$ individuals from the variation range of $\bm{\alpha}$. According to Eq. ~(\ref{eq5}), the evolution of SP is to obtain better and better solutions for individuals in EP, i.e. to minimize:
\begin{equation}
\begin{split}
\underset{\bm{\alpha}_{j}\in EP} {\textrm{max}} \ & (F(SP,\bm{\alpha}_{j})-\underset{\bm{x}}{\textrm{min}}f(\bm{x},\bm{\alpha}_{j})) \\
\textrm{where} \ & F(SP,\bm{\alpha}_{j})= \underset{\bm{x}\in SP} {\textrm{min}} f(\bm{x},\bm{\alpha}_{j})\\
\end{split}
\end{equation}
The evolution of EP is to obtain more and more challenging environment $\bm{\alpha}_{j}$ for individuals in SP, i.e. to maximize:
\begin{equation}
\begin{split}
& F(SP,\bm{\alpha}_{j})-\underset{\bm{x}}{\textrm{min}}f(\bm{x},\bm{\alpha}_{j}) \\
\textrm{where} \ & F(SP,\bm{\alpha}_{j})= \underset{\bm{x}\in SP} {\textrm{min}} f(\bm{x},\bm{\alpha}_{j})\\
\end{split}
\end{equation}
The SP at the last generation (i.e. $\mathit{SP}_{G_\mathit{max}}$) is the solution set $\bm{X}$ to find. It should be noted here $\underset{\bm{x}}{\textrm{min}}f(\bm{x},\bm{\alpha}_{j})$ is not known beforehand for optimization problems, thus other replacements of fitness evaluation should be used in the implementation of the proposed framework. For constraint satisfaction problems, $\underset{\bm{x}}{\textrm{min}}f(\bm{x},\bm{\alpha}_{j})$ can be set to zero. 

\begin{algorithm}[htbp]
	\algsetup{linenosize=\small} \small
	\caption{Co-evolutionary Framework for SP and EP \label{alg5}}
	\begin{algorithmic}[1]
		\REQUIRE Search space for the solution: $\bm{R}^{D_{x}}$, range for the enviromental parameters: $\bm{R}^{D_{a}}$, maximum number of generations: $G_{\mathit{max}}$ \\
		\STATE Set $g=0$
		\STATE Initialize solution population $\mathit{SP}_{0}$ based on $\bm{R}^{D_{x}}$ 
		\STATE Initialize environment population $EP_{0}$ based on $\bm{R}^{D_{a}}$
		\WHILE {$g\leq G_\mathit{max}$}
		\STATE Evolve $\mathit{SP}_{g}$ to get $\mathit{SP}_{g+1}$ based on $\mathit{EP}_{g}$
		\STATE Evolve $\mathit{EP}_{g}$ for one generation based on $\mathit{SP}_{g+1}$
		\STATE Set $g=g+1$
		\ENDWHILE
		\ENSURE Output $\mathit{SP}_{G_\mathit{max}}$
	\end{algorithmic}
\end{algorithm}

\subsubsection{Online Optimization}
After the system design phase ends, the solution set is obtained and then it is used to optimize the corresponding DCOP online. In the online optimization phase, the obtained solution set is used as the initial population at the beginning and each time a problem change is detected. During the period between two consecutive changes, local search operations are conducted on each individual in the population to address the current problem. This is to enable fast re-adaptation and parallel execution of local search on individuals to satisfy real-time requirements of some real-world applications. 

\subsubsection{Udpate of Solution Set}
Although the use of CC aims to explore as many different environmental changes as possible, it might miss some environmental changes due to limited computing resources. Therefore, the solution set found by CC offline should be updated online according to the true environmental changes. One potential way to update the solution set includes adding newly encountered environmental changes to the environment population $\mathit{EP}$ and running CC online. Another way is to add  the best solution found for each newly encountered environmental change to the solution set. 

More intuitively, the proposed framework suggests that EAs are used in the system design phase to obtain a component of a system for DOPs, while EAs are not necessarily (although they could be) be used as the solver explicitly once the system design is accomplished. Although such a framework might induce higher total computational cost in total (i.e., computational cost in the system design phase and computational cost for solving DOPs online), the additional cost could bring critical benefits. To be specific, by investing computational cost to obtain an archive of solutions in advance, the computational cost for solving DOPs online could be significantly alleviated. Such a benefit allows the system to response to the environmental change and output high quality solutions in a much faster (ideally in real-time) manner, which is an essential requirement for most solver systems designed for DOPs. Besides, the re-formulation of DOPs to static set-oriented optimization problems makes it possible to leverage the power of EAs to the largest extent, because the set-oriented optimization problem is much harder than its corresponding static version of the DOP of interest, and EAs typically show more advantages over traditional methods on set-oriented optimization, e.g., multi-objective optimization. 

\subsection{An Instantiation for Dynamic Constrained Optimization Problems}





The proposed dynamic optimization framework might not achieve an ideal effect for any scenario. Therefore, the framework needs to be instantiated and tested on concrete scenarios. In this paper, we think that DCOPs can be a suitable test bed for the general idea because:
\begin{enumerate}
	\item DCOPs, compared to unconstrained dynamic optimization, are of more practical significance (because most real-world problems have constraints) while less investigated. 
	\item For real-world DCOPs, it is often more important to get a feasible solution quickly, which in many cases enlarge the ``basin of attraction``. In other words, it is more likely that the advantage of the general idea could be more visible on DCOP, if there is any advantage. 
\end{enumerate}
Considering these, we further propose an instantiation of the general framework for DCOPs. The resultant algorithm is called CC for DCOPs (CCDO). The aim is to show the potential of the general idea by testing it on some important and typical problem scenario, i.e. DCOPs. 


Without loss of generality, the DCOPs considered in this paper has the following formulation:
\begin{equation}\label{eq6}
\begin{split}
\underset{\bm{x}} {\textrm{minimise}}\ \ \ & f(\bm{x},\bm{\alpha}(t)) \\
\textrm{s.t.} \ \ \ & g_{i}(\bm{x},\bm{\alpha}(t))\leq 0, i=1,2,...,k  \\
\textrm{s.t.} \ \ \ & h_{j}(\bm{x},\bm{\alpha}(t))=0, j=1,2,...,l    \\
\end{split}
\end{equation}
where $\bm{x}=[x_{0},x_{1},...,x_{D_{x}}]^{T}$ denotes the decision variable vector and $D_{x}$ is the dimension of the search space, $k$ and $l$ denotes the number of inequality constraints and equality constraints, $t$ denotes the time, and $\bm{\alpha}$ is the vector of environmental parameters which vary at a certain frequency as time goes by. The variation range of each environmental parameter in $\bm{\alpha}$ is assumed to be known beforehand in this paper.


\subsubsection{Offline Set Search  for DCOPs}
The offline set search in CCDO for DCOPs follows the framework in Algorithm~\ref{alg2}. Algorithm~\ref{alg3} gives the evolutionary process of SP for one generation. At every generation $g$, two individuals $\bm{x}_{1}$ and $\bm{x}_{2}$ are randomly selected from SP at the current generation (i.e. $\mathit{SP}_{g}$). Then, a new individual $\bm{x}_\mathit{new}$ is generated through evolutionary operation on $\bm{x}_{1}$ and $\bm{x}_{2}$ and added to $\mathit{SP}_{g}$ to get $\mathit{SP'}_{g}$. After this, the worst individual $\bm{x}_\mathit{worst}$ that has the smallest fitness will be deleted from $\mathit{SP'}_{g}$ and the remaining individuals in $\mathit{SP'}_{g}$ will enter the next generation as $\mathit{SP}_{g+1}$. 

In CCDO, the fitness of $\bm{x}_{i}$ in SP is defined as the performance drop of SP on EP after deleting $\bm{x}_{i}$ from SP. Suppose $F(\mathit{SP},\bm{\alpha})$ denotes the performance of SP for a dynamic problem under the environment $\bm{\alpha}$. It is calculated as the fitness of $\bm{x}_{\mathit{best}}$ that performs best on $\bm{\alpha}$ among all $\bm{x}_{i}$ in SP. If $\bm{x}_{\mathit{best}}$ is feasible for the problem under the environment $\bm{\alpha}$, then $F(\mathit{SP},\bm{\alpha})=f(\bm{x}_{\mathit{best}},\bm{\alpha})$; otherwise, $F(\mathit{SP},\bm{\alpha})$ is set to the sum of constraint violation values of $\bm{x}_{\mathit{best}}$ on $\bm{\alpha}$. Considering the problem scaling, we define the fitness of $\bm{x}_{i}$ in SP as the number of $\bm{\alpha}$ in EP on which the performance changes if deleting $\bm{x}_{i}$ from SP. That is:
\begin{equation}
Fit(\bm{x}_{i}) = \sum_{\bm{\alpha}\in \mathit{EP}}[F(\mathit{SP},\bm{\alpha})\neq F(\mathit{SP}/\bm{x}_{i},\bm{\alpha})]
\end{equation}
When comparing two individuals in SP that have the same $Fit(\bm{x}_{i})$, one individual is randomly selected. 

\begin{algorithm}[htbp]
	\algsetup{linenosize=\small} \small
	\caption{SP Evolution for One Generation \label{alg3}}
	\begin{algorithmic}[1]
		\REQUIRE Solution pop: $\mathit{SP}_{g}$, environment pop: $\mathit{EP}_{g}$\\
		\STATE Evaluate the fitness of each $\bm{x}$ in $\mathit{SP}_{g}$ based on $\mathit{EP}_{g}$
		\STATE Randomly select $\bm{x}_{1}$ and $\bm{x}_{2}$ from $\mathit{SP}_{g}$
		\STATE Generate $\bm{x}_\mathit{new}$ based on $\bm{x}_{1}$ and $\bm{x}_{2}$
		\STATE Set $\mathit{SP'}_{g}=\mathit{SP}_{g}\cup\bm{x}_\mathit{new}$
		\STATE Evaluate the fitness of each $\bm{x}$ in $\mathit{SP'}_{g}$
		\STATE Set $\mathit{SP}_{g+1}=\mathit{SP'}_{g}/\bm{x}_\mathit{worst}$
	\end{algorithmic}
\end{algorithm}

Algorithm~\ref{alg4} details the evolutionary process of EP for one generation. At every generation $g$, a new population $\mathit{EP}_\mathit{new}$ is generated based on $\mathit{EP}_{g}$ using evolutionary operation. Then, $\mathit{EP}_{g}$ and $\mathit{EP}_\mathit{new}$ are combined to obtain $\mathit{EP'}_{g}$. After this, $\mathit{EP}_{g}$ and $\mathit{EP}_\mathit{new}$ will be compared using a pair-wise comparison based on their challenging degree. All winners will enter next generation as $\mathit{EP}_{g+1}$. 

\begin{algorithm}[htbp]
	\algsetup{linenosize=\small} \small
	\caption{EP Evolution for One Generation \label{alg4}}
	\begin{algorithmic}[1]
		\REQUIRE Environment pop: $\mathit{EP}_{g}$, solution pop: $\mathit{SP}_{g}$\\
		\STATE Generate a new population $\mathit{EP}_\mathit{new}$ based on $\mathit{EP}_{g}$ using crossover and mutation
		\STATE Set $\mathit{EP'}_{g}=\mathit{EP}_{g}\cup \mathit{EP}_\mathit{new}$
		\STATE Evaluate each $\alpha$ in $\mathit{EP'}_{g}$
		\FOR {i=1,2,...,$\mathit{size\_of\_EP}$}
		\IF {$\mathit{EP}_{g}^{i}$ is more challenging than $\mathit{EP}_\mathit{new}^{i}$}
		\STATE $\mathit{EP}_{g+1}^{i} = \mathit{EP}_{g}^{i}$
		\STATE $\mathit{EP}_\mathit{worse}^{i}=\mathit{EP}_\mathit{new}^{i}$
		\ELSE 
		\STATE $\mathit{EP}_{g+1}^{i} = \mathit{EP}_\mathit{new}^{i}$
		\STATE $\mathit{EP}_\mathit{worse}^{i}=\mathit{EP}_{g}^{i}$
		\ENDIF
		\ENDFOR
	\end{algorithmic}
\end{algorithm}

An environment $\bm{\alpha}$ can be said to be more challenging if individuals in SP perform worse under this environment. Thus, to evaluate the challenging degree of an individual $\bm{\alpha}_{j}$ in EP, CCDO randomly generates a set of solutions (denoted as $I$) and uses the improvement obtained by SP over $I$ on $\bm{\alpha}_{j}$ to calculate the fitness. According to whether the best solutions for $\bm{\alpha}_{j}$ in $I$ and SP are feasible and whether an improvement is investigated, an individual $\bm{\alpha}_{j}$ can be categorised into the following four cases:
\begin{enumerate}
	\item The best solution for $\bm{\alpha}_{j}$ in SP is infeasible. In this case, the fitness of $\bm{\alpha}_{j}$ is defined as the sum of constraint violation values of the best solution.
	\item The best solution for $\bm{\alpha}_{j}$ in SP is feasible, but it is worse than the best solution for $\bm{\alpha}_{j}$ in $I$. In this case, the improvement obtained by $I$ over SP on $\bm{\alpha}_{j}$ is used to evaluate it. That is, 
	\begin{equation}
	Fit(\bm{\alpha}_{j})= \frac{\textrm{min}_{\bm{x}\in SP}F(x,\alpha_{j})-\textrm{min}_{\bm{x}\in I}F(x,\alpha_{j})}{\textrm{max}(|\textrm{min}_{\bm{x}\in I}F(x,\alpha_{j})|,|\textrm{min}_{\bm{x}\in SP}F(x,\alpha_{j})|)}
	\end{equation}
	The larger the improvement is, the more challenging $\bm{\alpha}_{j}$ is. 
	\item The best solution for $\bm{\alpha}_{j}$ in SP is feasible and the best solution in $I$ is infeasible. In this case, as the improvement is hard to calculate, the fitness of $\bm{\alpha}_{j}$ is defined as the function value of the best solution in SP for $\bm{\alpha}_{j}$ (i.e. $\textrm{min}_{\bm{x}\in \mathit{SP}}F(x,\alpha_{j}$)). 
	\item Both the best solutions for $\bm{\alpha}_{j}$ in SP and $I$ are feasible but the best solution in SP is better than the best solution in I. $\bm{\alpha}_{j}$ is considered less challenging if a larger improvement is obtained by SP over $I$. Thus, the fitness of $\bm{\alpha}_{j}$ is defined as the minus of the improvement obtained by SP over $I$ on $\bm{\alpha}_{j}$. That is, 
	\begin{equation}
	Fit(\bm{\alpha}_{j})= \frac{\textrm{min}_{\bm{x}\in SP}F(x,\alpha_{j})-\textrm{min}_{\bm{x}\in I}F(x,\alpha_{j})}{\textrm{max}(|\textrm{min}_{\bm{x}\in I}F(x,\alpha_{j})|,|\textrm{min}_{\bm{x}\in SP}F(x,\alpha_{j})|)}
	\end{equation} 
\end{enumerate}
In each case, the larger the fitness is, the more challenging $\bm{\alpha}_{j}$ is. When comparing individuals belonging to different cases, an individual belonging to case 1 is considered more challenging than an individual that belongs to case 2, an individual belonging to case 2 is considered more challenging than an individual belonging to case 3 or case 4. When comparing an individual belonging to case 3 and an individual belonging to case 4, a random one is selected. 

\subsubsection{Online Optimisation for DCOPs}
In the online optimisation process, CCDO reacts to truly happening changes of a DCOP based on the solution set ($\mathit{SP}_{G_\mathit{max}}$) obtained in the offline stage. Algorithm~\ref{alg2} gives the process of online optimisation. At the beginning, CCDO uses $\mathit{SP}_{G_\mathit{max}}$ to initialise a population $\mathit{pop}$. Then, at each generation, CCDO conducts local search on each individual in the population to do fast optimisation. In the filed of EAs, local search is usually used in memetic algroithms/computing \cite{neri2007adaptive, tirronen2008enhanced} to do exploitation in the neighborhood of an individual. 

\begin{algorithm}[htbp]
	\algsetup{linenosize=\small} \small
	\caption{Online Optimisation Process \label{alg2}}
	\begin{algorithmic}[1]
		\REQUIRE Search space for the solution: $\bm{R}^{D_{x}}$, number of individuals to detect changes: $\mathit{detect}_{k}$, $\mathit{SP}_{G_\mathit{max}}$ obtained from offline set search \\
		\STATE Initialise a population $\mathit{pop}$ based on $\mathit{SP}_{G_\mathit{max}}$
		\STATE Evaluate $\mathit{pop}$ with the current $f$
		\WHILE{stopping criteria is not satisfied}
		\FOR{each solution $\bm{x}_{i}$ in the population $\mathit{pop}$}
		\STATE Re-evaluate $\bm{x}_{i}$ to detect whether a change happens
		\IF{a change is detected}
		\STATE Update the solution set $\mathit{SP}_{G_\mathit{max}}$
		\STATE Re-initialise the population based on $\mathit{SP}_{G_\mathit{max}}\cup \mathit{LS}_\mathit{best}$
		\STATE Go to step 3
		\ELSE
		\IF{the distance between $\bm{x}_{i}$ and $mem_{ls}$ is larger than 1e-2}
		\STATE Do SQP local search on $\bm{x}_{i}$ to get $\bm{x}_{i}^{best}$
		\STATE Archive $\bm{x}_{i}$ into $mem_{ls}$
		\STATE Archive $\bm{x}_{i}^{best}$ into $\mathit{LS}_{best}$
		\STATE Set $\bm{x}_{i}=\bm{x}_{i}^{best}$
		\ENDIF
		\STATE Generate $\bm{x}'_{i}=\bm{x}_{i}+\delta*randn(1,D_{x})$
		\IF{$f(\bm{x}'_{i})$ is better than $f(\bm{x}_{i})$}
		\STATE Set $\bm{x}_{i}=\bm{x}'_{i}$
		\ENDIF
		\ENDIF
		\ENDFOR
		\STATE Re-evaluate $\mathit{detect_{k}}$ sentinel solutions to detect changes
		\IF{change is detected}
		\STATE Update the solution set $\mathit{SP}_{G_\mathit{max}}$
		\STATE Re-initialise the population $\mathit{pop}$ based on $\mathit{SP}_{G_\mathit{max}}\cup \mathit{LS}_\mathit{best}$
		\ENDIF
		\ENDWHILE	
	\end{algorithmic}
\end{algorithm}

In this paper, we implemented the local search operation by considering both the traditional sequential quadratic programming (SQP) \cite{boggs1995sequential} and the Gaussian mutation. To implement SQP, the Matlab optimisation toolbox and \textit{fmincon} function \cite{matlab2017} were used in this work. In SQP, the gradient information for the objective and constraint functions is estimated by the forward difference formula. More implementation details can be found in \cite{matlab2017}. After doing SQP, a Gaussian mutation is conducted on the resultant individual. The mutant individual will replace the original individual if it is better. To avoid doing SQP on similar individuals, individuals that have undergone SQP will be recorded in a memory, $mem_{ls}$. At every generation, if the distance between one individual and the nearest individual in the $mem_{ls}$ is less than 1e-2, SQP is omitted on this individual. At the end of each generation, individuals whose distance to the nearest individual in the $mem_{ls}$ is less than 1e-2 are replaced by random individuals. 

To detect changes, CCDO re-evaluates each individual in the current population before doing local search operation on it. Moreover, CCDO generates several random individuals and uses them as sentinels. After local search operation, CCDO re-evaluates sentinels to detect changes. If a change is detected, CCDO will first update the solution set $\mathit{SP}_{G_\mathit{max}}$ with the best solution found for last environmental change, and then re-initialise the population based on the updated $\mathit{SP}_{G_\mathit{max}}$ and the optimal solutions $\mathit{LS}_\mathit{best}$ obtained by the last local search operations. 

\section{Experimental Study}
In the experimental study, we aim to answer the following two questions:

\begin{enumerate}[{1.}]
	\item whether CCDO can significantly speed up the optimization of a DCOP?
	\item How is about the trade-off between the offline computational cost and the online optimization speed?
	\item Whether the online update of solution set can significantly improve the optimization?
\end{enumerate}

To answer the first question, a DCOP benchmark test set with different change frequencies was used to test the performance of CCDO and performance comparisons were made between it and existing methods. We also replaced some aspects of CCDO with other strategies to check the performance change on the test functions. To answer the second question, we compared the solution set found offline by CC and the solution set obtained by the method with a randomly generated set of environments. The following parts in this section will present experimental details and results. 

\subsection{Comparison between CCDO and Existing Methods}
The proposed CCDO method was compared with 4 state-of-the-art algorithms, SELS \cite{lu2016speciated}, DDECV \cite{ameca2014differential}, EBBPSO-T \cite{campos2015entropy}, LTFR-DSPSO \cite{bu2017continuous}. Different from SELS, DDECV and EBBPSO-T, LTFR-DSPSO applied a gradient-based repair method \cite{chootinan2006constraint} to handle the constraints but the cost of all feasibility calculations was ignored in LTFR-DSPSO. To make a fair comparison, the CCDO method was modified by using the same gradient-based repair method when compared to LTFR-DSPSO. The resultant method is recorded as CCDO+Repair. In CCDO+Repair, only the feasible solution and the sentinel solutions are evaluated with the objective function. For each infeasible solution, a repair operation is conducted. If the solution obtained from local search is still infeasible, the repair operation is also conducted on this solution. In this section, when comparing two different methods, the Wilcoxon rank-sum test with a confidence level at 0.05 was used. 

\subsubsection{Benchmark Problems}
In the literature, there exist two suites of DCOP benchmark test functions, one was proposed in \cite{nguyen2012continuous} and the other was given in \cite{bu2017continuous}. The latter one comprises test functions that have smaller feasible regions. Considering that DCOPs with small feasible regions usually prefer methods that deal with constraints using a repair mechanism, we only used the DCOP test problems in the former suite in this study. In the experiments, 9 DCOP benchmark test functions given in \cite{nguyen2012continuous} were used. These are G24-l (dF, fC), G24-2 (dF, fC), G24-3 (dF, dC), G24-3b (dF, dC), G24-4 (dF, dC), G24-5 (dF, dC), G24-6a (dF, fC), G24-6c (dF, fC), G24-7 (fF, dC). Here, `dF', `fF', `dC', and `fC' mean dynamic objective function, fixed objective function, dynamic constraint functions, and fixed constraint functions, respectively. All of these test functions are minimising problems. More details about the 9 test functions can be found in the Appendix. 

In our experiments, 6 settings for the change frequency of each test function were considered. They are 1000, 500, 250, 100, 50 and 25 objective function evaluations (FEs). Note that the setting of 100, 50 or 25 FEs means the problem changes very fast. As far as we know, none of the previous studies in the field of DCOPs have considered such settings. The change severity was set to be medium (i.e., $k=0.5$ and $s=20$). The environmental parameters for the 9 DCOP test functions are the variables in them that changes as time goes by. The details of the environmental parameters and their ranges in each test function are given in Table~\ref{tab1}. The range for each environmental parameter was set as the interval between the minimum value and the maximum value that the parameter can have when the number of problem changes is set to 12. 

An ideal solution set for a DCOP should be the one that comprises an optimal solution for any environment in the range of environmental parameter values. To better characterise the DCOP test functions, Table~\ref{tab1} gives the smallest size that an ideal solution set can have for each test function in the column of `solution size'. It can be seen from Table~\ref{tab1} that the 9 test functions can be classified into two groups. The first group includes G24-l (dF, fC), G24-2 (dF, fC), G24-6a (dF, fC), G24-6c (dF, fC), for which the smallest ideal solution set has a limited size. The second group includes G24-3 (dF, dC), G24-3b (dF, dC), G24-4 (dF, dC), G24-5 (dF, dC), G24-6d (dF, fC), and G24-7 (fF, dC), for which the smallest ideal solution set has an unlimited size. 

\begin{table}[!htbp]
	\begin{center}
		\caption{Environmental parameters for each test function \label{tab1}}
		\begin{tabular}{|c|c|c|c|}
			\hline
			Function & Parameters & Ranges & Solution Size\\
			\hline
			G24-l (dF,fC) & $p_{1}$ & [-1,1] & 2 \\
			\hline
			G24-2 (dF,fC) & $p_{1}$, $p_{2}$ & [-1,1],[-1,1] & 5 \\
			\hline
			G24-6a (dF,fC) &$p_{1}$ & [-1,1] & 2 \\			
			\hline
			G24-6c (dF,fC) &$p_{1}$ & [-1,1] & 2 \\
			\hline
			G24-3 (dF,dC) & $s_{2}$ & [-0.2,2] & many\\
			\hline
			G24-3b (dF,dC) & $p_{1}$, $s_{2}$ & [-1,1], [-0.2,2] & many\\
			\hline
			G24-4 (dF,dC) & $p_{1}$, $s_{2}$ & [-1,1], [0,2.2] & many \\
			\hline
			G24-5 (dF,dC) & $p_{1}$, $p_{2}$, $s_{2}$ & [-1,1],[-1,1],[0,2.2] & many \\
			\hline
			G24-7 (fF,dC) & $s_{2}$ & [0,2.2] & many \\
			\hline
		\end{tabular}
	\end{center}
\end{table}

\subsubsection{Performance Metrics}
As different performance metrics were considered to assess the efficacy of the 4 methods for comparison in their original papers, different performance metrics were also applied in this paper when making different comparisons. When comparing the CCDO with SELS, EBBPSO-T and DDECV, the modified offline error averaged at each function evaluation was used to evaluate the performance of each algorithm. The modified offline error averaged at each function evaluation is defined as follows:
\begin{equation}
E_{MO}=\frac{1}{num\_of\_eval}\sum_{j=1}^{num\_of\_eval}e_{MO}(j)
\end{equation}
where $num\_of\_eval$ denotes the maximum number of the function evaluations, and $e_{MO}(j)$ denotes the error of the best feasible solution obtained at $j$-th evaluation. If there are no feasible solutions at the $j$-th evaluation, the worst possible value that a feasible solution can have will be taken. The error value of a feasible solution means the difference between its function value and the best possible value that a feasible solution can have. The best and worst possible values that a feasible solution can have were approximated by experiments for each test function. The smaller the modified offline error is, the better the algorithm performs. 

When comparing the CCDO+Repair with LTFR-DSPSO, the modified offline error averaged at every generation was used to evaluate the performance of each algorithm. The modified offline error averaged at every generation is defined as follows:
\begin{equation}
E_{MO}=\frac{1}{num\_of\_gen}\sum_{j=1}^{num\_of\_gen}e_{MO}(j)
\end{equation}
where $num\_of\_gen$ denotes the maximum number of the function evaluations, and $e_{MO}(j)$ denotes the error of the best feasible solution obtained at $j$-th generation. If there are no feasible solutions at the $j$-th generation, the worst possible value that a feasible solution can have will be taken. 

\subsubsection{Parameter Settings}
When compared to SELS, EBBPSO-T and DDECV, the number of changes was set to 12, which is the same as in \cite{lu2016speciated}. The parameter settings for the competitive co-evolutionary search process and online optimisation are given in Tables ~\ref{tab2} and~\ref{tab3}, respectively. For the parameter setting of SQP, the other parameters were set as default except the parameters mentioned in Table~\ref{tab3}. The parameter setting for SELS is the same as in its original paper \cite{lu2016speciated}. The experimental results of EBBPSO-T and DDECV in their original papers were used for comparison. 

  \begin{table}[!htbp]
	\begin{center}
		\caption{Parameter settings for co-evolutionary search process when comparing CCDO with SELS, EBBPSO-T and DDECV \label{tab2}}
		\begin{tabularx}{0.43\textwidth}{|c|*{3}{Y|}}
			\hline
			$G_\mathit{max}$ &  50  & $\mathit{SP}_\mathit{size}$ & 10 \\
			\hline
			$\mathit{EAr}_\mathit{size}$ & 10  & $\mathit{EP}_\mathit{size}$ &  10 \\
			\hline
			$i_\mathit{size}$ &  5  & $\mathit{esp}$ &  50 \\
			\hline
			\multirow{2}{*}{SP evolution} & \multicolumn{3}{c|}{Gaussian mutation with scale = 0.1, rate = 0.5, } \\
			& \multicolumn{3}{c|}{Intermediate Crossover with rate = 0.5} \\
			\hline
			EP evolution & \multicolumn{3}{c|}{Gaussian mutation with scale = 0.05, rate = 0.5}\\
			\hline
		\end{tabularx}
	\end{center}
\end{table}

\begin{table}[!htbp]
	\begin{center}
		\caption{Parameter settings for online optimisation when comparing CCDO with SELS, EBBPSO-T and DDECV \label{tab3}}
		\begin{tabular}{|c|c|}
			\hline
			Parameter & Value \\
			\hline
			$detect_{k}$ & 4 \\
			\hline
			\multirow{3}{*}{SQP} & ConstraintTolerance = 0 \\
			& HonorBounds = true \\
			& MaxFunctionEvaluations = 20 \\
			\hline
            Mutation & Gaussian mutation with scale = 0.1\\
			\hline
		\end{tabular}
	\end{center}
\end{table}

When compared to LTFR-DSPSO, the number of changes was set to 10, which is the same as in \cite{bu2017continuous}. The parameter settings for the competitive co-evolutionary search process and online optimisation are given in Table~\ref{tab4} and Table~\ref{tab5}, respectively. For the parameter setting of SQP, the other parameters were set as default except for the parameters mentioned in Table~\ref{tab5}. The experimental results of LTFR-DSPSO given in their original papers were used for comparison. 

\begin{table}[!htbp]
	\begin{center}
		\caption{Parameter settings for co-evolutionary search process when comparing CCDO with LTFR-DSPSO \label{tab4}}
		\begin{tabularx}{0.43\textwidth}{|c|*{3}{Y|}}
			\hline
			$G_{max}$ & 50 & $SP_{size}$ & 20 \\
			\hline
			$EAr_{size}$ & 20 & $EP_{size}$ & 20 \\
			\hline
			$i_{size}$ & 5 & $esp$ & 50 \\
			\hline
			\multirow{2}{*}{SP evolution} & \multicolumn{3}{c|}{Gaussian mutation with scale = 0.1, rate = 0.5,} \\
			& \multicolumn{3}{c|}{Intermediate Crossover with rate = 0.5} \\
			\hline
			EP evolution &  \multicolumn{3}{c|}{Gaussian mutation with scale = 0.05, rate = 0.5}\\
			\hline
		\end{tabularx}
	\end{center}
\end{table}

\begin{table}[!htbp]
	\begin{center}
		\caption{Parameter settings for online optimisation when comparing CCDO with LTFR-DSPSO \label{tab5}}
		\begin{tabular}{|c|c|}
			\hline
			Parameter & Value \\
			\hline
			$detect_{k}$ & 4 \\
			\hline
			\multirow{3}{*}{SQP} & ConstraintTolerance = 0 \\
			& HonorBounds = true \\
			& MaxFunctionEvaluations = 50 \\
			\hline
			Mutation & Gaussian mutation with scale = 0.1\\
			\hline
		\end{tabular}
	\end{center}
\end{table}

\subsubsection{Comparison Results under Change Frequency of 1000 FEs}
The most commonly used setting of change frequency in the previous studies is 1000 FEs. Table~\ref{tab7} summarises the mean and standard deviation of the modified offline error over 50 runs obtained by DDECV, EBBPSO-T, SELS and the proposed CCDO method under a change frequency of 1000 FEs. In Table~\ref{tab7}, the best result obtained on each test function is marked in bold according to statistical test. 

It can be seen from Table~\ref{tab7} that CCDO generally performed best and SELS performed second best among the 4 algorithms. According to the `dF', `fF', `dC', and `fC' features of test functions, the 9 test functions can be classified into three groups: (dF, fC), (fF, dC) and (dF, dC). It can be observed that CCDO performed better than other methods on the groups of (dF, fC) and (fF, dC) but worse on the group of (dF, dC). For the test functions in the group of (dF, dC), both the objective function and constraint functions change over time. This means the number of possible environments in this group is bigger than in other groups, which might pose a higher requirement of the solution set. In future work, CCDO will be further improved by introducing random individuals in the initial population in addition to the found solution set and using other evolutionary operations in addition to local search.

\begin{table}[!htbp]
	\begin{center}
		\caption{Comparison results between DDECV, EBBPSO-T, SELS and the CCDO method under a change frequency of 1000 FEs. Better results obtained on each test function are marked in \textbf{bold}. \label{tab7}}
		\resizebox{0.5\textwidth}{!}{
			\begin{tabular}{|c|c|c|c|c|}
				\hline
				Func & DDECV & EBBPSO-T & SELS & CCDO \\
				\hline
				G24-l (dF,fC) & 0.109$\pm$0.033 & 0.084$\pm$0.041 & 0.025$\pm$0.008 & \textbf{0.009$\pm$0.008} \\
				\hline
				G24-2 (dF,fC) & 0.126$\pm$0.030 & 0.136$\pm$0.013 & 0.050$\pm$0.015 & \textbf{0.015$\pm$0.006}\\
				\hline
				G24-3 (fF,dC) & 0.057$\pm$0.018  & 0.032$\pm$0.005 & 0.044$\pm$0.022 & \textbf{0.015$\pm$0.007} \\
				\hline
				G24-3b (dF,dC) & 0.134$\pm$0.033 & 0.104$\pm$0.015 & 0.052$\pm$0.018 & \textbf{0.039$\pm$0.028} \\
				\hline
				G24-4 (dF,dC) & 0.131$\pm$0.032 & 0.138$\pm$0.022 & \textbf{0.082$\pm$0.021} & 0.121 $\pm$0.030 \\
				\hline
				G24-5 (dF,dC) & 0.126$\pm$0.030 & 0.126$\pm$0.019 &  \textbf{0.054$\pm$0.014} & 0.083$\pm$0.021 \\
				\hline
				G24-6a (dF,fC) & 0.215$\pm$0.067 & 0.116$\pm$0.099  & 0.055$\pm$0.009 & \textbf{0.033$\pm$0.008} \\
				\hline
				G24-6c (dF,fC) & 0.128$\pm$0.025 & 0.251$\pm$0.061 & 0.052$\pm$0.008 &  \textbf{0.031$\pm$0.007} \\
				\hline
				G24-7 (fF,dC)  & 0.106$\pm$0.022 & \textbf{0.045$\pm$0.009} & 0.087$\pm$0.016 & 0.083$\pm$0.013 \\
				\hline
			\end{tabular}
		}
	\end{center}
\end{table}

In Table~\ref{tab8}, we also summarise the mean and standard deviation of the modified offline error over 50 runs obtained by LTFR-DSPSO and the proposed CCDO under a change frequency of 1000 FEs. The best result obtained on each test function is marked in bold. It can be seen from Table~\ref{tab8} that the CCDO+Repair achieved better results than LTFR-DSPSO except for only three test functions. 

\begin{table}[!htbp]
	\begin{center}
		\caption{Comparison results between the LTFR-DSPSO method and the CCDO+Repair method. Better results obtained on each test function are marked in \textbf{bold}. \label{tab8}}
		\resizebox{0.43\textwidth}{!}{
		\begin{tabular}{|c|c|c|}
			\hline
			Func & LTFR-DSPSO & CCDO+Repair \\
			\hline
			G24-l (dF,fC) & 6.09e$-$06$\pm$4.24e$-$05 & \textbf{7.09e$-$08$\pm$1.96e$-$07} \\
			\hline
			G24-2 (dF,fC) & 8.54e$-$04$\pm$3.38e$-$03 & \textbf{6.15e$-$04$\pm$3.07e$-$03} \\
			\hline
			G24-3 (fF,dC) & 3.64e$-$05$\pm$1.13e$-$04 & \textbf{7.87e$-$11$\pm$3.04e$-$10} \\
			\hline
			G24-3b (dF,dC) & 3.82e$-$05$\pm$1.17e$-$04 & \textbf{7.80e$-$08$\pm$3.78e$-$07} \\
			\hline
			G24-4 (dF,dC) & \textbf{5.45e$-$06$\pm$3.79e$-$05} & 1.81e$-$02$\pm$9.07e$-$02 \\
			\hline
			G24-5 (dF,dC) & 7.00e$-$05$\pm$4.89e$-$04 & \textbf{9.92e$-$07$\pm$4.94e$-$06} \\
			\hline
			G24-6a (dF,fC) & \textbf{2.21e$-$18$\pm$9.28e$-$18} & 5.91e$-$04$\pm$9.80e$-$04 \\
			\hline
			G24-6c (dF,fC) & \textbf{1.91e$-$18$\pm$8.82e$-$18} & 6.39e$-$04$\pm$1.35e$-$03 \\
			\hline
			G24-7 (fF,dC)   & 5.11e$-$06$\pm$3.55e$-$05 &  \textbf{1.28e$-$13$\pm$1.17e$-$13} \\
			\hline
		\end{tabular}
	}
	\end{center}
\end{table}

\subsubsection{Comparison Results under Change Frequency of 500 FEs, 250 FEs, 100 FEs, 50 FEs and 25 FEs}
To evaluate the performance of CCDO on fast-changing DCOPs, experiments were also conducted under five other change frequencies in addition to 1000 FEs. As no complete results were found for EBBPSO-T, DDEVC and LTFR-DSPSO on the 9 test functions under these change frequencies, CCDO was compared in this experiment only to SELS which performed the second best under a change frequency of 1000 FEs. Tables~\ref{tab9} and~\ref{tab10} give the comparison results between SELS and CCDO under change frequencies of 500 FEs, 250 FEs, 100 FEs, 50 FEs and 25 FEs, respectively. 

It can be seen by comparing Tables~\ref{tab7} and~\ref{tab9} that the advantage of CCDO over SELS is similar when the change frequency changes from 1000 FEs to 500 FEs or 250 FEs. However, when the change frequency is further reduced from 1000 FEs to 100 FEs, 50 FEs or 25 FEs, it can be seen from the Tables~\ref{tab7} and~\ref{tab10} that the advantage of CCDO over SELS becomes more obvious. For each change frequency of 100 FEs, 50 FEs and 25 FEs, CCDO obtained better results than SELS on all 9 test functions. 

Fig.~\ref{fig2} give the evolutionary curves for both CCDO and SELS on one representative test function, G24-l (dF, fC). The evolutionary curves for the other test functions can be found in Fig. 1-8 in the Appendix. It can be seen that SELS did not work properly when the problem changes fast (e.g. under the change frequency of 50 FEs or 25 FEs). In contrast, CCDO still showed a fast adaptation under such settings. All these results demonstrate the advantage of CCDO in solving fast-changing DCOPs. 

\subsubsection{Performance Analysis of CCDO}
We further conducted experiments to analyse the effect of the solution set and SQP on the performance of CCDO. In this experiment, we ran CCDO with randomly generated individuals for population initialisation and re-initialisation instead of using the solution set. The resultant method is recorded as CCDO-S. We also ran CCDO with another local search operation, local evolutionary search enhancement by random memorizing (LESRM) \cite{voigt1998local} used in SELS \cite{lu2016speciated}, instead of SQP. The resultant method is recorded as CCDO-L. Tables~\ref{tab17} gives the comparison results between CCDO-S and CCDO under change frequency of 100 FEs. 

It can be seen that the performance deteriorated a lot without using the solution set. Note that the comparison results under other change frequencies are similar. This demonstrates the benefits of offline set search. Tables~\ref{tab18} gives the comparison results between CCDO-L and CCDO under change frequencies of 100 FEs, respectively. We can observe that the use of SQP is slightly better than the use of the other local search. Moreover, when comparing CCDO-L to SELS, we found that CCDO-L still outperformed SELS. Thus, LESRM can be used as a replacement when the Matlab toolbox of SQP is not available. 

\begin{table}[!htbp]
	\begin{center}
		\caption{Comparison results between CCDO and CCDO-S. Better results obtained on each test function are marked in \textbf{bold}. \label{tab17}}
		\resizebox{0.43\textwidth}{!}{
			\begin{tabular}{|c|c|c|}
				\hline
				Func & CCDO-S & CCDO\\
				\hline
				G24-l (dF,fC) & 6.14e$-$01$\pm$2.80e$-$01 & \textbf{1.10e$-$01$\pm$1.08e$-$01}  \\
				\hline
				G24-2 (dF,fC) & 5.79e$-$01$\pm$1.18e$-$01 & \textbf{1.42e$-$01$\pm$4.23e$-$02} \\
				\hline
				G24-3 (fF,dC) & 7.57e$-$01$\pm$1.47e$-$01 &  \textbf{1.51e$-$01$\pm$6.97e$-$02} \\
				\hline
				G24-3b (dF,dC) & 1.08e$+$00$\pm$1.65e$-$01 & \textbf{4.91e$-$01$\pm$1.72e$-$01}  \\
				\hline
				G24-4 (dF,dC) & 1.41e$+$00$\pm$2.60e$-$01 &  \textbf{7.49e$-$01$\pm$1.23e$-$01}  \\
				\hline
				G24-5 (dF,dC) & 9.33e$-$01$\pm$1.63e$-$01 & \textbf{4.73e$-$01$\pm$1.12e$-$01}  \\
				\hline
				G24-6a (dF,fC) & 1.87$+$00$\pm$2.01e$-$01 & \textbf{2.64e$-$01$\pm$9.16e$-$02}  \\
				\hline
				G24-6c (dF,fC) & 1.77e$+$00$\pm$1.65e$-$01 & \textbf{3.05e$-$01$\pm$9.40e$-$02} \\
				\hline
				G24-7 (fF,dC)   & 1.14e$+$00$\pm$2.45e$-$01 &  \textbf{6.43e$-$01$\pm$1.10e$-$01} \\
				\hline
			\end{tabular}
		}
	\end{center}
\end{table}

\begin{table}[!htbp]
	\begin{center}
		\caption{Comparison results between CCDO and CCDO-L. Better results obtained on each test function are marked in \textbf{bold}. \label{tab18}}
		\resizebox{0.43\textwidth}{!}{
			\begin{tabular}{|c|c|c|}
				\hline
				Func & CCDO-L & CCDO\\
				\hline
				G24-l (dF,fC) & \textbf{8.41e$-$02$\pm$6.87e$-$02} & 1.10e$-$01$\pm$1.08e$-$01  \\
				\hline
				G24-2 (dF,fC) &1.30e$-$01$\pm$3.78e$-$02 & 1.42e$-$01$\pm$4.23e$-$02 \\
				\hline
				G24-3 (fF,dC) & 2.05e$-$01$\pm$6.42e$-$02 &  \textbf{1.51e$-$01$\pm$6.97e$-$02} \\
				\hline
				G24-3b (dF,dC) & 5.68e$-$01$\pm$1.56e$-$01 & \textbf{4.91e$-$01$\pm$1.72e$-$01}  \\
				\hline
				G24-4 (dF,dC) & 7.77e$-$01$\pm$1.26e$-$01 &  7.49e$-$01$\pm$1.23e$-$01  \\
				\hline
				G24-5 (dF,dC) & 4.65e$-$01$\pm$1.03e$-$01 & 4.73e$-$01$\pm$1.12e$-$01 \\
				\hline
				G24-6a (dF,fC) & 4.58e$-$01$\pm$2.78e$-$02 & \textbf{2.64e$-$01$\pm$9.16e$-$02}  \\
				\hline
				G24-6c (dF,fC) & 3.97e$-$01$\pm$6.31e$-$01 & \textbf{3.05e$-$01$\pm$9.40e$-$02} \\
				\hline
				G24-7 (fF,dC)   & 7.71e$-$01$\pm$8.73e$-$02 &  \textbf{6.43e$-$01$\pm$1.10e$-$01} \\
				\hline
			\end{tabular}
		}
	\end{center}
\end{table}

\begin{table*}[!htbp]
	\begin{center}
		\caption{Comparison results between SELS and the CCDO method under change frequencies of 500 FEs and 250 FEs, respectively. Better results obtained on each test function are marked in \textbf{bold}. \label{tab9}}
		\resizebox{0.70\textwidth}{!}{
			\begin{tabular}{|c|c|c|c|c|}
				\hline
				\multirow{2}{*}{Func} & \multicolumn{2}{c|}{500 FEs} & \multicolumn{2}{c|}{250 FEs} \\
				\cline{2-5}
				& SELS & CCDO & SELS & CCDO \\
				\hline
				G24-l (dF,fC) & 6.18e$-$02$\pm$1.65e$-$02 & \textbf{2.42e$-$02$\pm$4.11e$-$02} & 1.49e$-$01$\pm$4.91e$-$02 & \textbf{6.20e$-$02$\pm$1.05e$-$01} \\
				\hline
				G24-2 (dF,fC) & 8.98e$-$02$\pm$1.48e$-$02 & \textbf{3.14e$-$02$\pm$1.48e$-$02} & 2.02e$-$01$\pm$2.93e$-$02 & \textbf{6.09e$-$02$\pm$2.36e$-$02}  \\
				\hline
				G24-3 (fF,dC) & 8.75e$-$02$\pm$2.94e$-$02 & \textbf{3.00e$-$02$\pm$1.67e$-$02} & 1.85e$-$01$\pm$4.87e$-$02 & \textbf{5.99e$-$02$\pm$3.38e$-$02} \\
				\hline
				G24-3b (dF,dC) & 1.12e$-$01$\pm$2.86e$-$02 & \textbf{9.73e$-$02$\pm$7.24e$-$02} & \textbf{2.37e$-$01$\pm$4.51e$-$02} & \textbf{2.89e$-$01$\pm$1.95e$-$01} \\
				\hline
				G24-4 (dF,dC) &  \textbf{1.52e$-$01$\pm$3.70e$-$02} & 2.34e$-$01$\pm$9.20e$-$02 & \textbf{3.35e$-$01$\pm$5.78e$-$02} & 4.40e$-$01$\pm$1.51e$-$01 \\
				\hline
				G24-5 (dF,dC) & \textbf{9.90e$-$02$\pm$1.54e$-$02} & 1.51e$-$01$\pm$4.08e$-$02 & \textbf{2.16e$-$01$\pm$3.45e$-$02} & 2.67e$-$01$\pm$6.98e$-$02 \\
				\hline
				G24-6a (dF,fC) & 1.08e$-$01$\pm$1.45e$-$02 & \textbf{6.54e$-$02$\pm$1.30e$-$02} &  2.14e$-$01$\pm$4.13e$-$02 & \textbf{1.16e$-$01$\pm$3.84e$-$02} \\
				\hline
				G24-6c (dF,fC) & 1.09e$-$01$\pm$1.74e$-$02 & \textbf{6.28e$-$02$\pm$1.43e$-$02} & 1.96e$-$01$\pm$3.50e$-$02 & \textbf{1.11e$-$01$\pm$3.88e$-$02}  \\
				\hline
				G24-7 (fF,dC)  &  \textbf{1.36e$-$01$\pm$2.02e$-$02} & 1.48e$-$01$\pm$2.68e$-$02 & 4.08e$-$01$\pm$5.90e$-$02 & \textbf{2.96e$-$01$\pm$4.91e$-$02} \\
				\hline
			\end{tabular}
		}
	\end{center}
\end{table*} 

\begin{table*}[!htbp]
	\begin{center}
		\caption{Comparison results between SELS and the CCDO method under change frequencies of 100 FEs, 50 FEs and 25 FEs, respectively. Better results obtained on each test function are marked in \textbf{bold}. \label{tab10}}
		\resizebox{1.0\textwidth}{!}{
			\begin{tabular}{|c|c|c|c|c|c|c|c|c|}
				\hline
				\multirow{2}{*}{Func} & \multicolumn{2}{c|}{100 FEs} & \multicolumn{2}{c|}{50 FEs} & \multicolumn{2}{c|}{25 FEs} \\
				\cline{2-7}
				& SELS & CCDO & SELS & CCDO & SELS & CCDO \\
				\hline
				G24-l (dF,fC) & 4.05e$-$01$\pm$9.64e$-$02 & \textbf{1.10e$-$01$\pm$1.08e$-$01} 
				& 1.05e$+$00$\pm$2.17e$-$01 & \textbf{1.61e$-$01$\pm$1.09e$-$01} 
				& 1.55e$+$00$\pm$3.73e$-$01 & \textbf{2.69e$-$01$\pm$1.14e$-$01}  \\
				\hline
				G24-2 (dF,fC) & 3.96e$-$01$\pm$5.56e$-$02 & \textbf{1.42e$-$01$\pm$4.23e$-$02} 
				& 7.93e$-$01$\pm$1.12e$-$01 & \textbf{2.29e$-$01$\pm$6.24e$-$02} 
				& 9.91e$-$01$\pm$1.55e$-$01 & \textbf{3.75e$-$01$\pm$1.09e$-$01} \\
				\hline
				G24-3 (fF,dC) & 4.87e$-$01$\pm$1.35e$-$01 & \textbf{1.51e$-$01$\pm$6.97e$-$02}  
				& 9.76e$-$01$\pm$2.19e$-$01 & \textbf{2.92e$-$01$\pm$1.06e$-$01} 
				& 1.38e$+$00$\pm$2.82e$-$01 &  \textbf{4.41e$-$01$\pm$1.14e$-$01}  \\
				\hline
				G24-3b (dF,dC) & 5.43e$-$01$\pm$8.92e$-$02 & \textbf{4.91e$-$01$\pm$1.72e$-$01}  
				& 1.10e$+$00$\pm$1.52e$-$01 & \textbf{6.45e$-$01$\pm$1.82e$-$01} 
				& 1.63e$+$00$\pm$2.41e$-$01 & \textbf{8.61e$-$01$\pm$1.68e$-$01} \\
				\hline
				G24-4 (dF,dC) &  8.54e$-$01$\pm$1.29e$-$01 & \textbf{7.49e$-$01$\pm$1.23e$-$01} 
				& 1.50e$+$00$\pm$2.06e$-$01 & \textbf{9.83e$-$01$\pm$1.94e$-$01} 
				&  1.79e$+$00$\pm$2.82e$-$01 & \textbf{1.28e$+$00$\pm$2.16e$-$01}  \\
				\hline
				G24-5 (dF,dC) & 5.25e$-$01$\pm$8.99e$-$02 & \textbf{4.73e$-$01$\pm$1.12e$-$01} 
				& 1.03e$+$00$\pm$1.76e$-$01 & \textbf{6.50e$-$01$\pm$1.17e$-$01} 
				& 1.11e$+$00$\pm$1.90e$-$01 & \textbf{8.68e$-$01$\pm$1.27e$-$01} \\
				\hline
				G24-6a (dF,fC) & 5.53e$-$01$\pm$9.16e$-$02 & \textbf{2.64e$-$01$\pm$9.16e$-$02} 
				& 1.64e$+$00$\pm$2.72e$-$01 & \textbf{6.65e$-$01$\pm$2.42e$-$01} 
				& 2.17e$+$00$\pm$3.47e$-$01 & \textbf{1.00e$+$00$\pm$3.58e$-$01} \\
				\hline
				G24-6c (dF,fC) & 5.10e$-$01$\pm$7.52e$-$02 & \textbf{3.05e$-$01$\pm$9.40e$-$02}
				& 1.57e$+$00$\pm$2.86e$-$01 & \textbf{6.68e$-$01$\pm$2.25e$-$01}
				& 1.99e$+$00$\pm$3.73e$-$01 & \textbf{8.60e$-$01$\pm$3.92e$-$01 } \\
				\hline
				G24-7 (fF,dC)  & 9.43e$-$01$\pm$2.05e$-$01 & \textbf{6.43e$-$01$\pm$1.10e$-$01}
				& 1.83e$+$00$\pm$2.75e$-$01 & \textbf{1.03e$+$00$\pm$1.83e$-$01}
				& 2.07e$+$00$\pm$2.75e$-$01 & \textbf{1.55e$+$00$\pm$2.28e$-$01}  \\
				\hline
			\end{tabular}
		}
	\end{center}
\end{table*}

\subsection{The Effectiveness of Co-evolutionary Set Search}
In this experiment, we randomly generated the same number of environments at the beginning but fixed them during the co-evolutionary search process in Algorithm~\ref{alg5}. The solution set found by this fixed process was compared to the solution set found by the co-evolutionary search process. The parameter setting for these experiments was set according to Table~\ref{tab2}. 

To quantify the performance of the resulting solution sets, for each DCOP test function 50 sub-problems with randomly generated parameter values were generated as test sub-problems. The performance of the solution set on each sub-problem equals to the error value of the best solution in the solution set for this sub-problem. The error value of a feasible solution means the difference between its function value and the best possible value that a feasible solution can have. When there is no feasible solution in the solution set for a sub-problem, the worst possible objective function value that a feasible solution can have on this sub-problem is taken. The average performance of the solution set on these 50 sub-problems over 50 runs was used for comparison. Table~\ref{tab6} summarises the mean and standard deviation of the average best error obtained on 50 sub-problems over 50 runs for each test function. Better results are marked in bold according to the statistical test. 

\begin{table}[htbp]
	\begin{center}
		\caption{Comparison results between the co-evolutionary method and the random method with fixed environments. Better results obtained on each test function are marked in \textbf{bold}. \label{tab6}}
		\resizebox{0.47\textwidth}{!}{
			\begin{tabular}{|c|c|c|c|}
				\hline
				\multirow{2}{*}{Func} & Solution & 	\multirow{2}{*}{Fixed Environments} & \multirow{2}{*}{Co-evolutionary} \\
				& Size & & \\
				\hline
				G24-l (dF,fC) & 2 & \textbf{4.06e$-$02$\pm$6.25e$-$02} & 7.98e$-$02$\pm$1.15e$-$01 \\
				\hline
				G24-2 (dF,fC) & 5 &  \textbf{5.19e$-$02$\pm$8.57e$-$02} & 6.50e$-$02 $\pm$8.64e$-$02 \\
				\hline
				G24-6a (dF,fC) & 2 & \textbf{4.52e$-$02$\pm$1.90e$-$01} & 7.46e$-$02$\pm$2.63e$-$01 \\
				\hline
				G24-6c (dF,fC) & 2 & 4.45e$-$02$\pm$4.51e$-$02 & 3.38e$-$02$\pm$3.21e$-$02 \\
				\hline
				G24-3 (fF,dC) & many & 6.46e$-$01$\pm$5.47e$-$01 & \textbf{2.79e$-$01$\pm$3.14e$-$01} \\
				\hline
				G24-3b (dF,dC) & many & 6.03e$-$01$\pm$2.04e$-$01 & 5.85e$-$01$\pm$1.58e$-$01 \\
				\hline
				G24-4 (dF,dC) & many & 6.40e$-$01$\pm$2.14e$-$01 & 6.04e$-$01$\pm$1.75e$-$01 \\
				\hline
				G24-5 (dF,dC) & many & 2.86e$-$01$\pm$1.25e$-$01 & 2.81e$-$01$\pm$9.12e$-$02 \\
				\hline
				G24-7 (fF,dC) & many & 4.11e$-$01$\pm$1.97e$-$01 & \textbf{2.50e$-$01$\pm$6.42e$-$02} \\
				\hline
			\end{tabular}
		}
	\end{center}
\end{table}

It can be seen from Table~\ref{tab6} that the evolution of environments can help to obtain a competitive or better solution set on the second group of test functions in which the smallest ideal solution set has an unlimited size. But, on the first group of test functions in which the smallest ideal solution set has a limited size, the random method with fixed environments performed better. 

This is due to the fact that for the first group of test problems the coverage on all possible sub-problems can be easily achieved by random environment sampling since the smallest ideal solution set has a limited size. Hence, the evolution of environments can not bring too much improvement in locating a better solution set but needs more function evaluations. Consequently, the co-evolutionary method performs weaker than the one without the evolution of environments. 

In contrast, for the second group of test problems the coverage based on several randomly generated environments is rather weak, although the evolution of the environment population required more function evaluations. In this case, the evolution of environment population takes effect and thus the co-evolutionary method can show some advantages on this group of test problems. 

\section{Conclusions and Future Work}
\subsection{Conclusions}
In this paper, we proposed a new dynamic optimisation approach, CCDO, to address fast-changing DCOPs. It dynamically maintains a solution set which is obtained through CC at the beginning, and conducts online local search by using this set as initial solutions once a change has been detected. 

To evaluate the efficacy of CCDO, 9 DCOP benchmark test functions were used and 3 more change frequencies (100 FEs, 50 FEs and 25 FEs) that have not been considered before in the literature were used to test the performance of CCDO under fast changes. The experimental results demonstrated that CCDO still worked well on fast-changing DCOPs when SELS failed. We also conducted experiments to check the effectiveness of the environmental evolution in the co-evolutionary search process. The experimental results showed that the evolution of environments is more suitable for problems in which the smallest ideal solution set has an unlimited size compared to the random method with fixed environments. 

\subsection{Limitation of the Proposed Approach and Future Work}
The experimental study in this paper represents the first step to validate that the proposed CCDO approach can react rapidly to fast changes. In our future work, the CCDO method will be tested on more dynamic test problems and real-world applications. Note that the CCDO method can be generally applied to dynamic optimisation problems (dynamic constraint satisfaction problems, dynamic optimisation without constraints or with box constraints). In the experiments, the number of fitness evaluations needed in the offline set search process is 57150. As the cost in co-evolutionary search process was ignored in the experiments, in our future work, the performance of CCDO will be evaluated with the co-evolutionary search cost considered. Moreover, we will investigate improving CCDO by using other evolutionary operations and accelerating CCDO by parallel computing. 

Some open issues also arise from this work. First, for dynamic optimisation problems, it is important to investigate whether a solution set of limited size exists for which optimal solutions for any given environmental change can be found by doing local search on them. Second, the optimal size of a suitable solution set in the proposed dynamic optimisation framework needs to be studied. Third, whether and when CC is the most effective method to search a good solution set for a dynamic problem needs further investigation.


%



\section*{Acknowledgment}
This work was supported in part by the National Key R\&D Program of China (Grant No. 2017YFC0804003), the Program for Guangdong Introducing Innovative and Enterpreneurial Teams (Grant No. 2017ZT07X386), Shenzhen Peacock Plan (Grant No. KQTD2016112514355531), the Science and Technology Innovation Committee Foundation of Shenzhen (Grant No. ZDSYS201703031748284), EPSRC (Grant Nos. EP/J017515/1 and EP/P005578/1), and the Program for University Key Laboratory of Guangdong Province (Grant No. 2017KSYS008). The authors would like to thank Dr. Chenyang Bu for providing the source code for LTFR-DSPSO.

\ifCLASSOPTIONcaptionsoff
  \newpage
\fi



\bibliographystyle{IEEEtran}
\bibliography{IEEEabrv,mybib}
%

\newpage
\begin{figure*}
	\centering
	\subfigure[1000 FEs]{\includegraphics[width=0.45\textwidth]{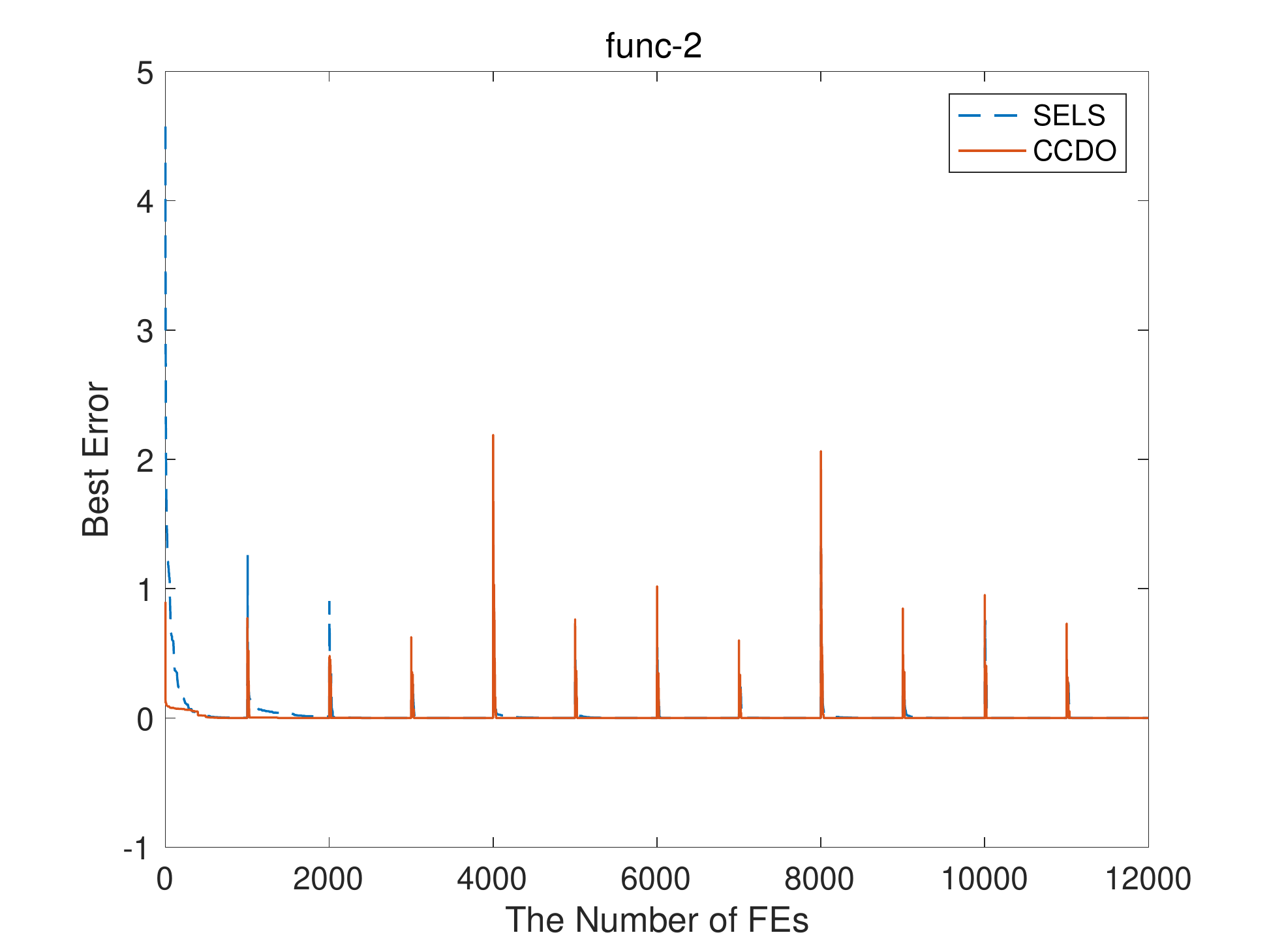}} 
	\subfigure[500 FEs]{\includegraphics[width=0.45\textwidth]{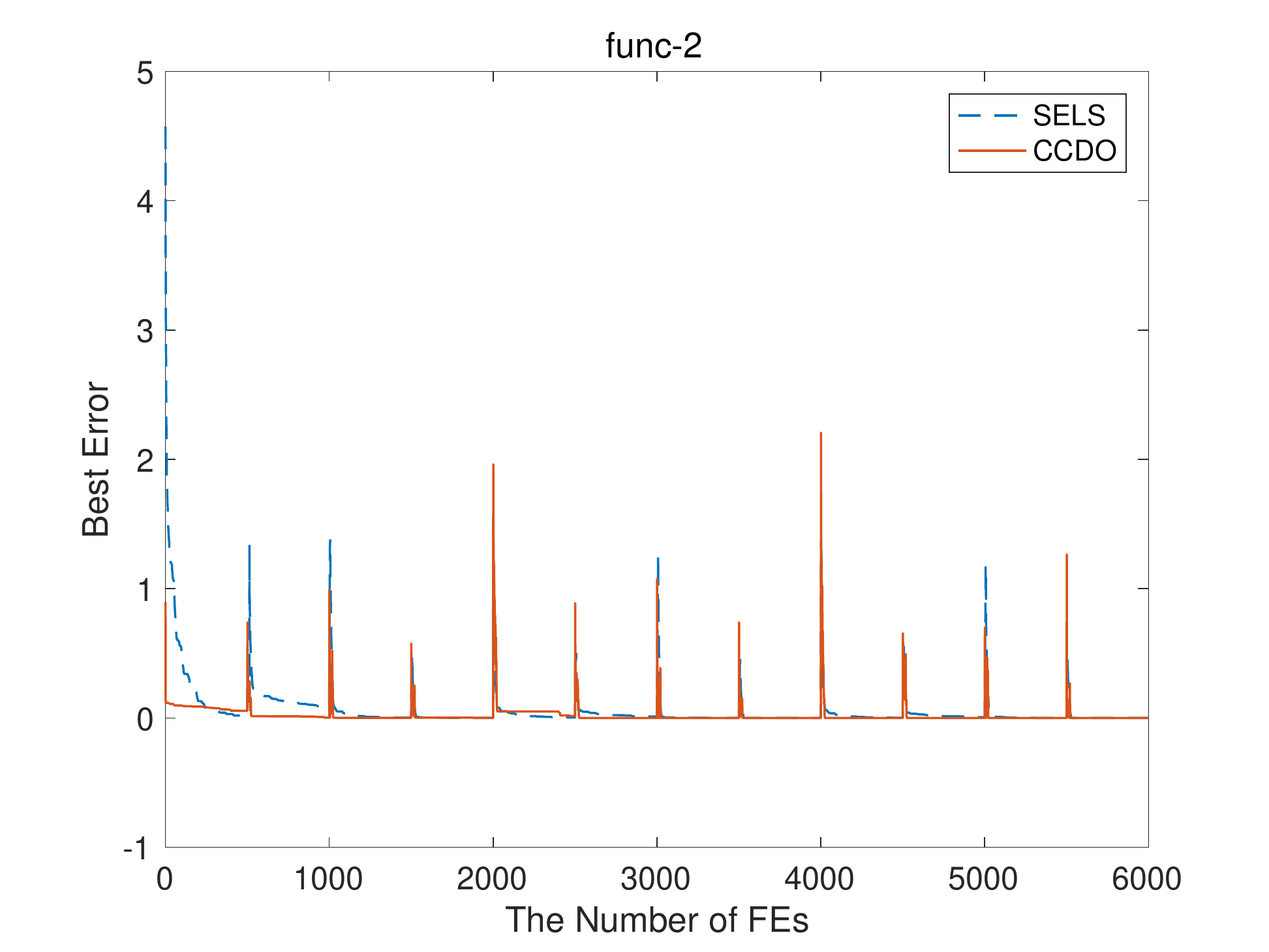}}
	\subfigure[250 FEs]{\includegraphics[width=0.45\textwidth]{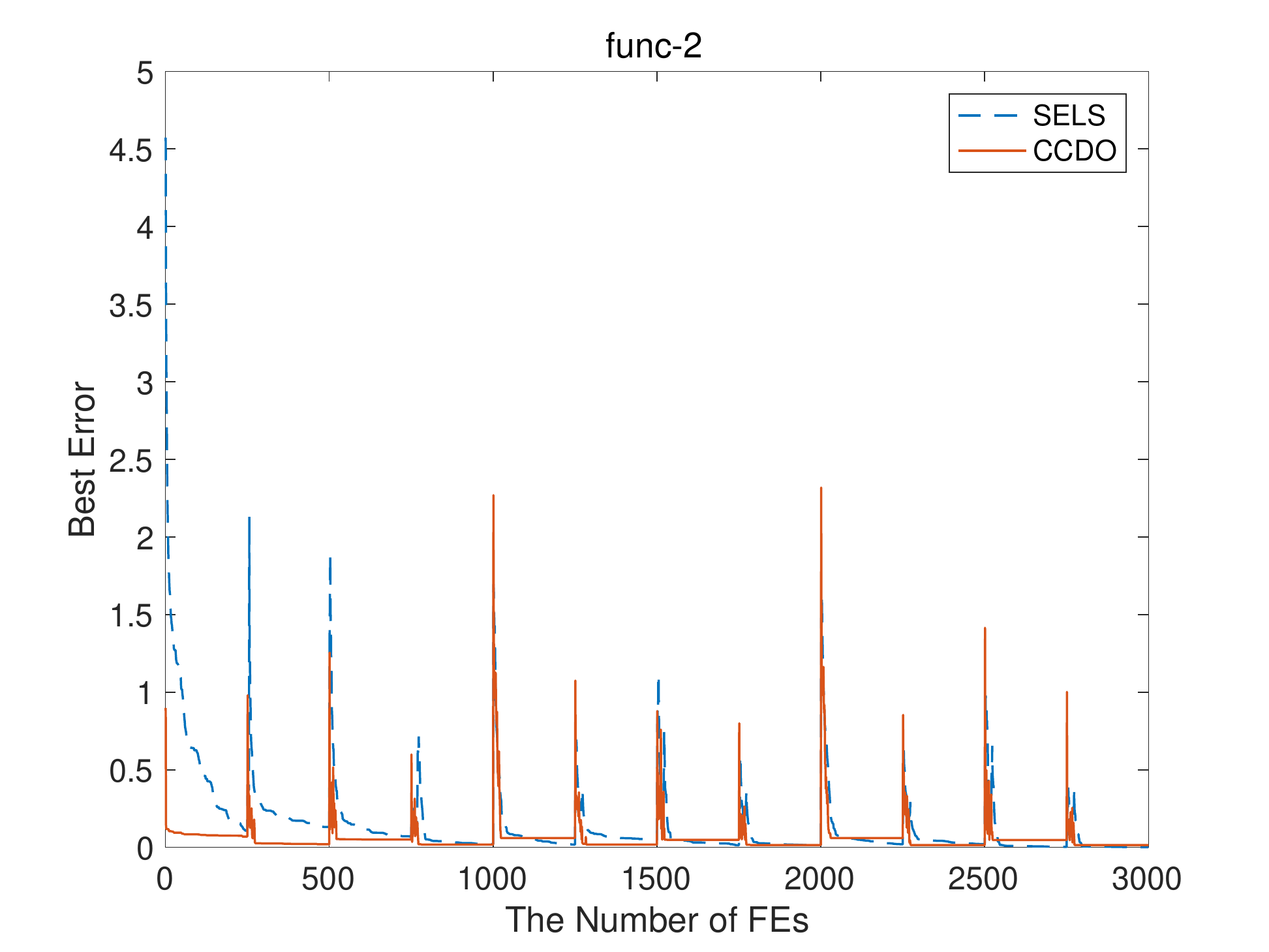}}  
	\subfigure[100 FEs]{\includegraphics[width=0.45\textwidth]{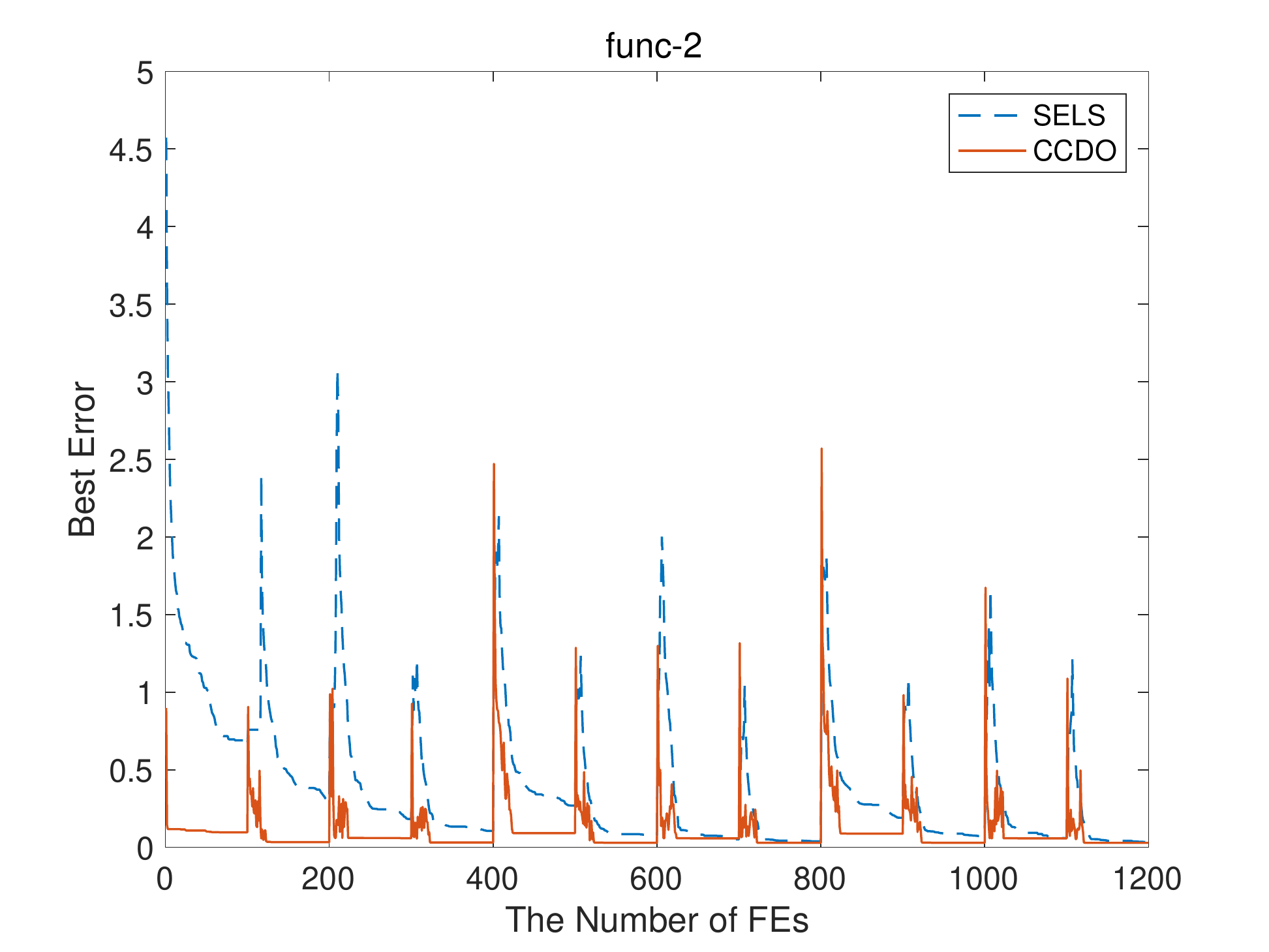}}  
	\subfigure[50 FEs]{\includegraphics[width=0.45\textwidth]{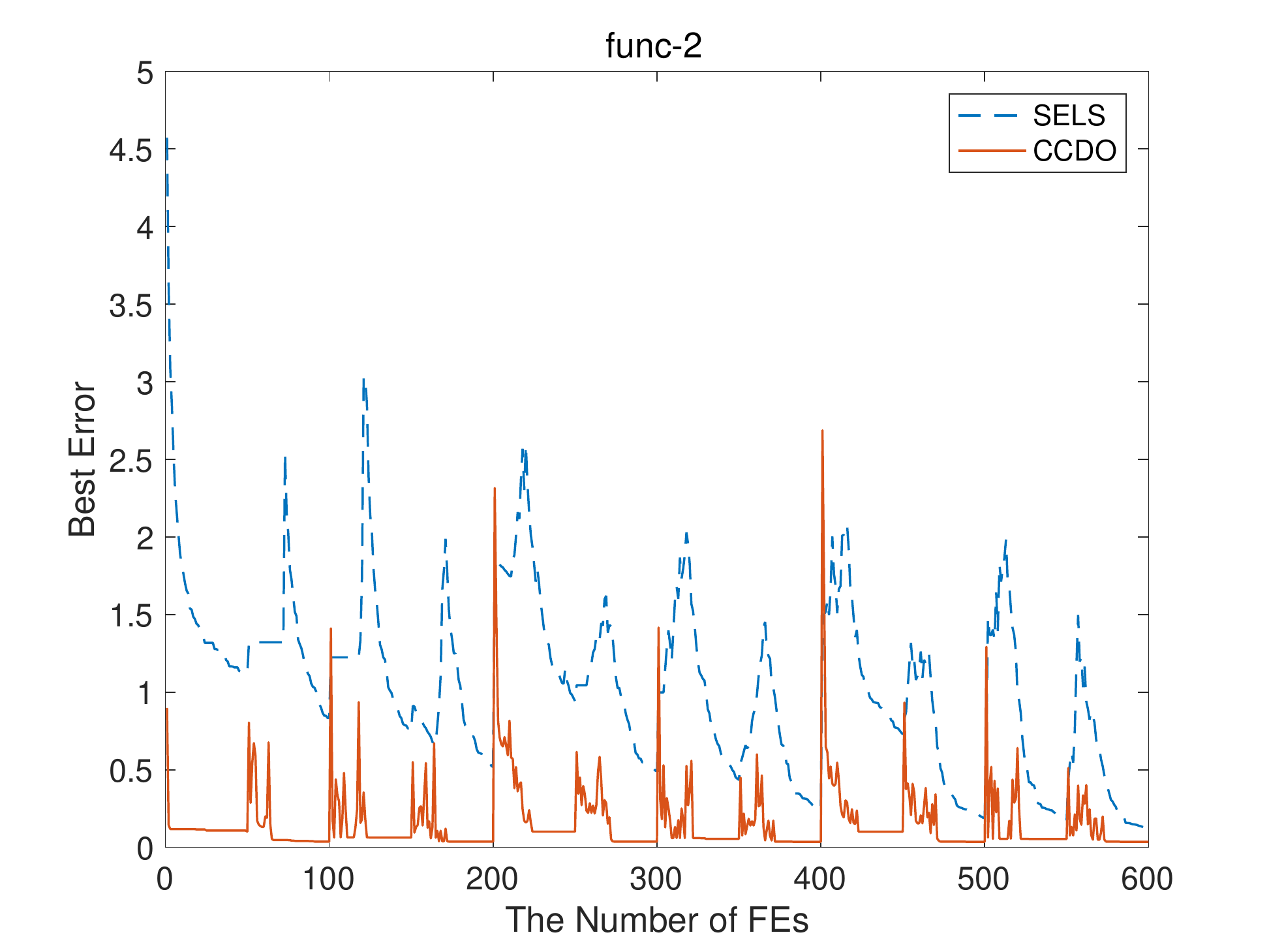}} 
	\subfigure[25 FEs]{\includegraphics[width=0.45\textwidth]{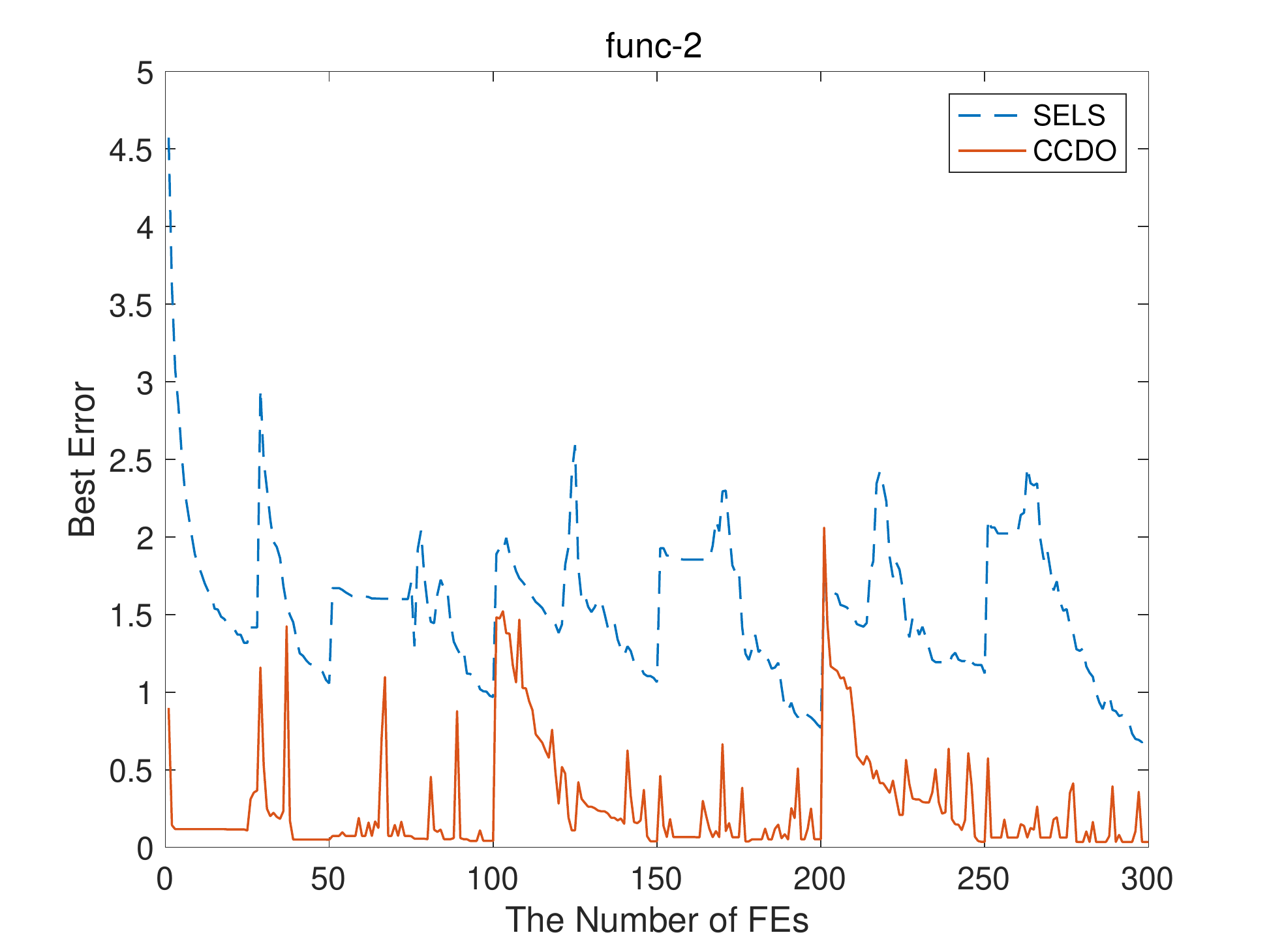}} 
	\caption{Evolutionary Curves of SELS and the CCDO method on G24-l (dF,fC) \label{fig3}}
\end{figure*}

%








\end{document}